\documentclass[letterpaper, 10pt, conference]{ieeeconf}
\IEEEoverridecommandlockouts  % to override locked commands
\usepackage{cite}
\usepackage{amsmath,amssymb,amsfonts}
\usepackage{algorithmic}
\usepackage{graphicx}
\usepackage{textcomp}
\usepackage{xcolor}

\usepackage{flushend}
\usepackage{multirow}
\usepackage{caption}
\usepackage{subcaption}
\usepackage{float}
\usepackage{booktabs}
\usepackage[capitalize]{cleveref}
\usepackage{dblfloatfix}

\usepackage{pifont}
\newcommand{\xmark}{\ding{55}}% creates the \xmark command
\usepackage{rotating} % for rotating figures
\usepackage{tabularx} % Include the tabularx package

\crefformat{figure}{Fig.\hspace{0.33em}#2#1#3}

\usepackage[absolute,showboxes]{textpos}

\TPMargin*{3pt}
\newcommand\copyrighttext{
    \footnotesize
    \noindent
    SUBMITTED TO REVIEW AND POSSIBLE PUBLICATION. COPYRIGHT WILL BE TRANSFERRED WITHOUT NOTICE.\\
    Personal use of this material is permitted.
    Permission must be obtained for all other uses, in any current or future media, including reprinting/republishing this material for advertising or promotional purposes, creating new collective works, for resale or redistribution to servers or lists, or reuse of any copyrighted component of this work in other works.}%
\newcommand\copyrightnotice{%
    \begin{textblock*}{6.6in}(0.95in,0.15in)
        \centering
        \copyrighttext
    \end{textblock*}
}

\begin{document}
\copyrightnotice

\title{\LARGE \bf Beyond Features: How Dataset Design Influences \\Multi-Agent Trajectory Prediction Performance}

\author{Tobias Demmler$^{1}$, Jakob Häringer$^{2}$, Andreas Tamke$^{3}$, Thao Dang$^{4}$, Alexander Hegai$^{5}$, and Lars Mikelsons$^{6}$% <-this % stops a space
\thanks{$^{1}$Tobias Demmler is with Robert Bosch GmbH, Stuttgart, Germany
        {\tt\small tobias.demmler@de.bosch.com}}%
\thanks{$^{2}$Jakob Häringer is with Robert Bosch GmbH, Stuttgart, Germany
        {\tt\small jakob.haeringer@de.bosch.com}}%
\thanks{$^{3}$Andreas Tamke is with Robert Bosch GmbH, Stuttgart, Germany
        {\tt\small andreas.tamke@de.bosch.com}}%
\thanks{$^{4}$Thao Dang is with the Institute for Intelligent Systems, Department of Computer Science and Engineering, Esslingen University, Germany
        {\tt\small thao.dang@hs-esslingen.de}}%
\thanks{$^{5}$Alexander Hegai is with Robert Bosch GmbH, Stuttgart, Germany
        {\tt\small alexander.hegai@de.bosch.com}}%
\thanks{$^{6}$Lars Mikelsons leads the Chair of Mechatronics of the University of Augsburg, Germany
        {\tt\small lars.mikelsons@uni-a.de}}%
}

\maketitle

\begin{abstract}

Accurate trajectory prediction is critical for safe autonomous navigation, yet the impact of dataset design on model performance remains understudied. This work systematically examines how feature selection, cross-dataset transfer, and geographic diversity influence trajectory prediction accuracy in multi-agent settings. We evaluate a state-of-the-art model using our novel L4 Motion Forecasting dataset based on our own data recordings in Germany and the US. This includes enhanced map and agent features. We compare our dataset to the US-centric Argoverse~2 benchmark. First, we find that incorporating supplementary map and agent features unique to our dataset, yields no measurable improvement over baseline features, demonstrating that modern architectures do not need extensive feature sets for optimal performance. The limited features of public datasets are sufficient to capture convoluted interactions without added complexity. Second, we perform cross-dataset experiments to evaluate how effective domain knowledge can be transferred between datasets. Third, we group our dataset by country and check the knowledge transfer between different driving cultures. 

\end{abstract}

\section{Introduction}
\begin{table*}[!bp]
\vspace{-1.5mm}
\centering
\captionsetup{width=0.75\linewidth}
\caption[Comparison of L4 to Other Motion Forecasting Datasets]{Comparison between our L4 Motion Forecasting dataset and other motion forecasting datasets adapted from Table 2 in \cite{Argoverse2_Original} and Table 1 in \cite{WaymoOriginal}.}
\begin{tabular}{@{}l|cc|ccc|c@{}}
  \toprule 
  Dataset & \multicolumn{2}{c|}{Argoverse}
  & Waymo & Lyft \cite{Lyft} & Shifts & L4\\
  Version & v1.1 \cite{Argoverse1} & v2 \cite{Argoverse2_Original} & v1.1 \cite{WaymoOriginal} & & v1 \cite{ShiftsYandex} & (ours)\\ \midrule
  
  Scenarios & 324k & 250k & 104k & 170k & 600k & 90k \\
  Total Time & 320 h & 763 h & 574 h & 1118 h & 1667 h & 274 h\\
  Cities & 2 & 6 & 6 & 1 & 6 & 2\\
  Unique Roadways & 290 km & 2220 km & 1750 km & 10 km & - & 250 km\\
  Sampling Rate & 10 Hz & 10 Hz & 10 Hz & 10 Hz & 5 Hz & 10 Hz\\
  Scenario Duration & 5 s & 11 s & 9.1 s & 25 s & 10 s & 11 s\\
  Forecast Horizon & 3 s & 6 s & 8 s & 5 s & 5 s & 6 s\\
  Average Tracks Per Scenario & 50 & 73 & - & 79 & 29 & 48\\
  Average Track Length & 2.48 s & 5.16 s & 7.04 s & 1.8 s & - & 3.43 s \\
  Multi-agent Evaluation & \xmark & \checkmark & \checkmark & \checkmark & \checkmark & \checkmark\\
  Evaluated Object Categories & 1 & 5 & 3 & 3 & 2 & 5\\
  Mined For Interestingness & \checkmark & \checkmark & \checkmark & - & \xmark & \xmark \\
  Object Bounding Boxes & \xmark & \xmark & 3D & 2D & 3D & 2D \\
  HD Maps               & 3D & 3D & 3D & 2D & 2D & 3D\\
  Traffic Signal States & \xmark & \xmark & \checkmark & \checkmark & \checkmark & \xmark\\
  \bottomrule
\end{tabular}

\label{tab:l4_vs_other_datasets}
\end{table*}

Trajectory prediction is a crucial part of modern autonomous driving systems. It enables these systems to understand and anticipate the movement of other traffic participants. To learn these behavior patterns comprehensive datasets are required. In recent years, numerous datasets have been published. Each dataset exhibits distinct characteristics, particularly in focus and scope. Some datasets focus on urban driving, some on highway scenarios, or a mixture of both. The size of the datasets is also different for each of them. Some datasets are only recorded in a single city. Others are recorded in multiple cities but still only in a single country. The agent details can range from simple x,~and~y coordinates to detailed bounding boxes and object types. Another differentiation point is the included High Definition map (HD map). Notable datasets are for example the Argoverse datasets \cite{Argoverse1, Argoverse2_Original}, the Waymo Open Motion dataset \cite{WaymoOriginal}, and the Shifts dataset \cite{ShiftsYandex}. All these differences make it hard to evaluate which features are important for trajectory prediction and which features do not contribute to more detailed predictions. Additionally, all of these public datasets are most likely less detailed than the proprietary version used internally by these companies. 

\begin{figure}
    \vspace{1.5mm}
    \centering
    \includegraphics[width=0.95\columnwidth]{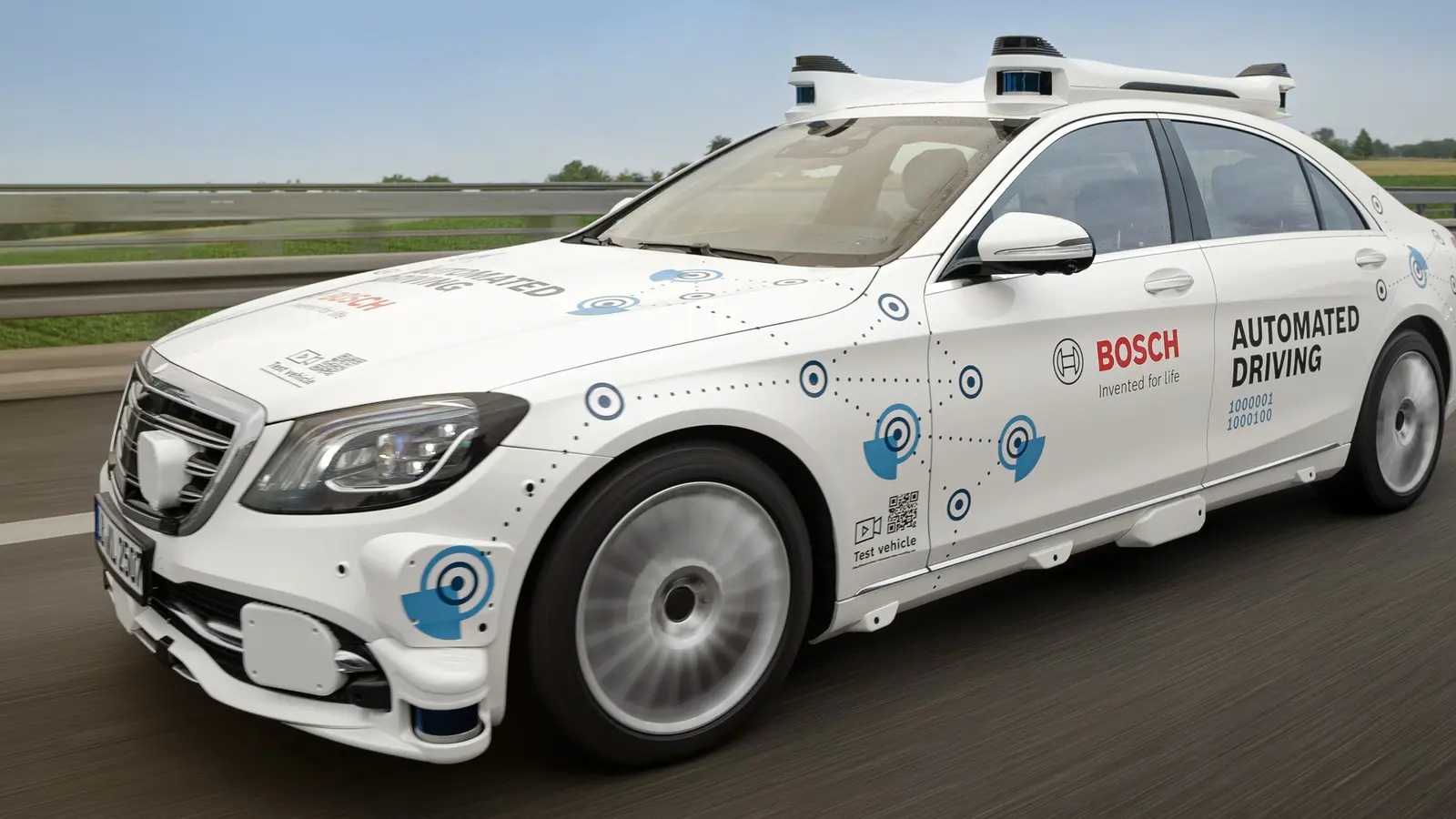}
    \caption{Test vehicle used to record data}
    \label{fig:test_vehicle}
    %\vspace{-5mm}
\end{figure}

Based on a selection of our own data recorded by our test vehicles (see \cref{fig:test_vehicle}), we created the L4 Motion Forecasting dataset. It contains data recorded in two countries, namely Germany and the United States. This allows us to evaluate how good motion forecasting models can transfer knowledge from one area of the world to another region. Additionally, we have access to all of the raw data. This enables us to evaluate whether using features not found in public datasets improves results. To do so, we use QCNet \cite{QCNet}, a state-of-the-art motion forecasting network at the top of the Argoverse~2 leaderboard \cite{Argoverse2Challenge}. We elected to construct the L4 Motion Forecasting dataset similar to the Argoverse 2 dataset. By using the same data structure as Argoverse~2 we can use our dataset with the public implementation of QCNet without adding modifications. Optionally, to leverage the additional features provided in our dataset only small modifications are required.

Thus, our contributions in this paper are twofold:
\begin{itemize}
    \item We explore our custom L4 Motion Forecasting dataset which differentiates itself from other datasets in location and included details.
    We evaluate these differences in detail and expect that our analysis can be transferred to other custom datasets.
    \item We train a state-of-the-art motion forecasting model on this dataset and provide an in-depth analysis of the results. Demonstrating that additional features are not required for state-of-the-art performance. Additionally, we investigate the effects of knowledge transfer between datasets and geographic regions.
\end{itemize}

\section{Related Work}

Over the past few years, datasets and benchmarks for perception tasks in autonomous driving have had a tremendous impact on the computer vision community. One of the earliest datasets, KITTI \cite{KITTI}, played a pivotal role in opening up new research directions and establishing foundational benchmarks. However, a significant limitation of early datasets like KITTI was their lack of integrated map data, despite the critical role that detailed maps play in developing real-world autonomous systems. Publicly available maps such as OpenStreetMap~\cite{OpenStreetMap}, while useful, lack the detail and accuracy required for advanced motion prediction tasks. 

This gap was first addressed by the introduction of the Argoverse 3D Tracking Dataset (Argoverse~1) \cite{Argoverse1}, which was the first to integrate high-definition (HD) maps into a large-scale motion forecasting benchmark. Chang~et~al.~\cite{Argoverse1} demonstrated that these detailed maps can improve the accuracy of motion prediction models by providing precise geometric and semantic information about the environment, including lane boundaries and crosswalks. This integration of HD maps marked a significant advancement in the field, setting a new standard for subsequent datasets.

Recent advancements in motion prediction for autonomous driving have been significantly influenced by the development of diverse datasets, each addressing specific research objectives and computational constraints. \cref{tab:l4_vs_other_datasets} highlights key datasets, including Argoverse~1 \cite{Argoverse1}, which introduced HD maps but is limited by its short forecast horizon and lack of multi-agent evaluation. Argoverse~2 (AV2) \cite{Argoverse2_Original} extends these capabilities with longer durations, diverse scenarios, and enhanced multi-agent interactions, though it lacks certain features like object bounding boxes and traffic signal states. While the dataset was extended to six different cities, it still only contains US cities. The same is true for the Waymo Open Motion Dataset \cite{WaymoOriginal, WaymoLidar}. It contains detailed 3D annotations but requires significant computational resources because of its size and the inclusion of raw lidar measurements. Lyft L5 \cite{Lyft} focuses on a specific route, providing high-resolution spatial detail but limited geographic applicability. Shifts \cite{ShiftsYandex} emphasizes robustness against out-of-distribution data, though it lacks detailed environmental features like 3D HD maps. Collectively, these datasets underscore the need for continued innovation in dataset development to address existing gaps and further advance motion prediction technologies. Our L4 Motion Forecasting dataset addresses multiple of these shortcomings. 

Parallel to the evolution of these public datasets neural networks leveraging the provided details were developed. VectorNet \cite{VectorNet} and LaneGCN \cite{LaneGCN} were early models to leverage HD maps. Many subsequent models used methods of these two approaches, such as TNT \cite{Zhao2020TNTTT} and HeteroGAT~\cite{TobiasPaper}. The dominance of transformer-based architectures in large language models also transferred to the motion forecasting domain with models such as Motion Transformers~(MTR)~\cite{MTR} and its derivatives \cite{shi2022mtra, ShiMTRV3, Demmler2025DynamicIQ} or QCNet \cite{QCNet} for example. 

For the evaluation of our dataset we use QCNet, a state-of-the-art transformer-based network achieving top results on the Argoverse 2 benchmark \cite{Argoverse2Challenge}. The codebase for this model is publicly available on GitHub \cite{QCNetGithub}.

\section{Dataset}
We base our experiments on our own L4 Motion Forecasting dataset. It was recorded in Germany and the United States. We structured it similarly to the AV2 dataset \cite{Argoverse2_Original} to allow for cross-compatibility.

\subsection{Data Collection}
\begin{figure}[ht]
    \vspace{1.5mm}
    \centering
    \subfloat[Stuttgart, Germany]{\includegraphics[width=0.8\columnwidth]{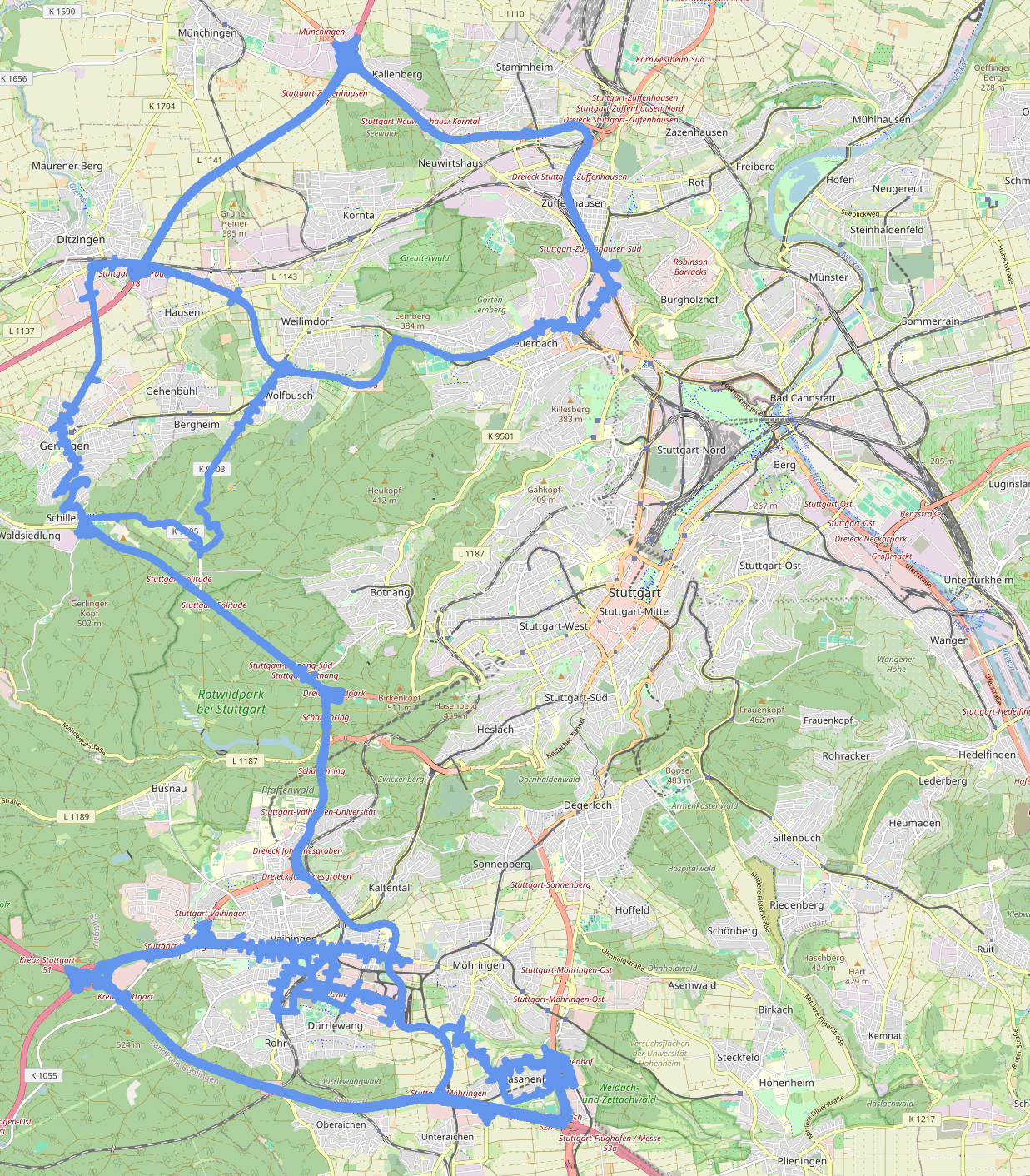}\label{fig:unique_roadways_stg}}
    \\\vspace{1mm}
    \subfloat[Sunnyvale, USA]{\includegraphics[width=\columnwidth]{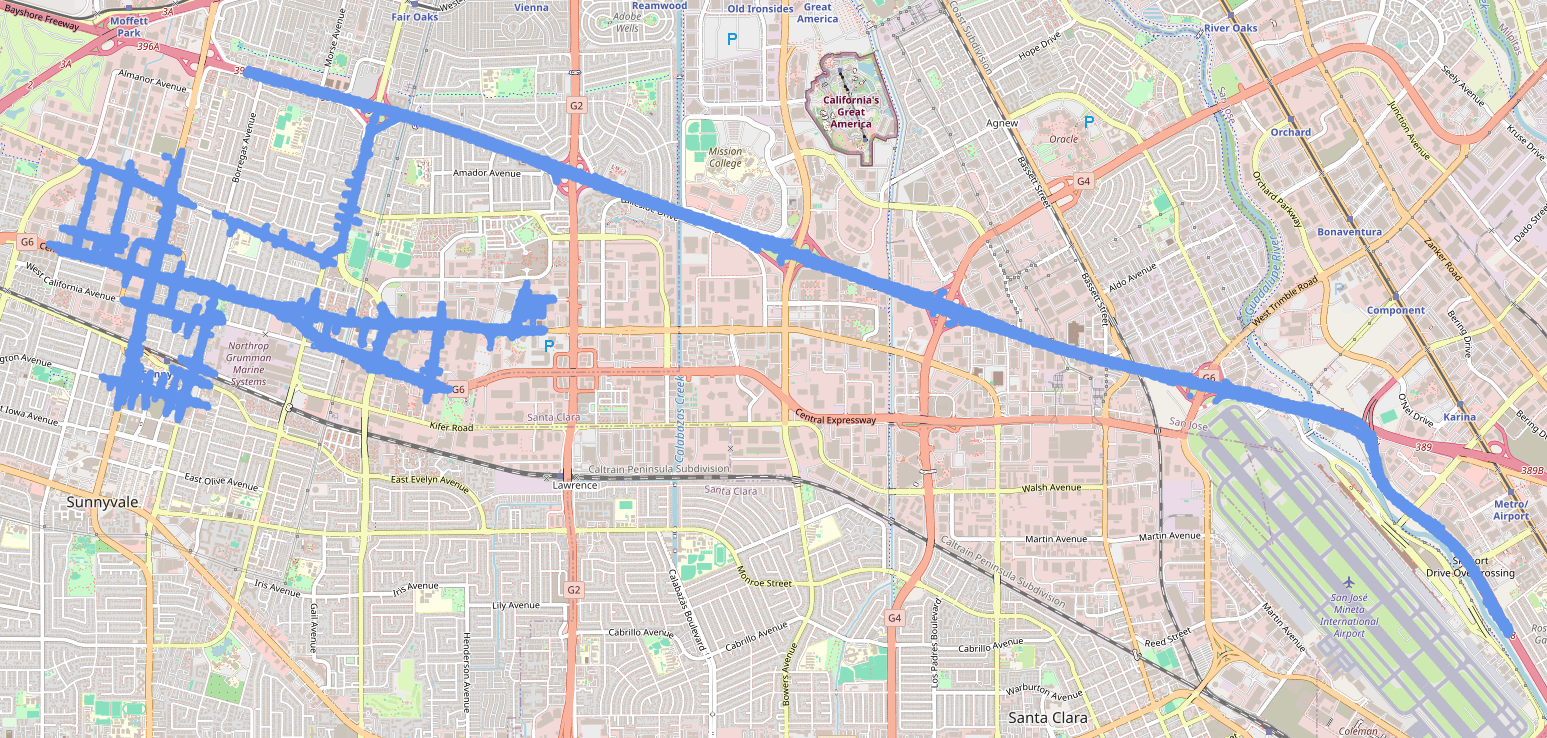}\label{fig:unique_roadways_sunnyvale}}
    \caption[]{Recording locations for the L4 Motion Forecasting dataset.}
    \label{fig:unique_roadways}
\end{figure}

The dataset is based on a selection of our own data which was recorded in Stuttgart (Germany) and Sunnyvale (United States) over two years starting in April 2022 until April 2024. Our test vehicles were equipped with a state-of-the-art sensor set including multiple cameras, lidars, and radars (see \cref{fig:test_vehicle}). The recorded locations include a wide variety of scenarios ranging from slow urban driving to fast driving on the German highway system which is famous for having sections without a speed limit. The precise locations of our recordings can be seen in \cref{fig:unique_roadways_stg} and \cref{fig:unique_roadways_sunnyvale}. In total over 400 hours of raw driving data were used for this dataset.

\subsection{Raw Data}\label{sec:data}

The raw data is recorded at a sampling rate of 10 Hz, which ensures that the positions of each agent are captured consistently over time. Alongside this positional data, each agent is represented by a bounding box that delineates its spatial dimensions. Additionally, the dataset includes several extra attributes for the agents, providing deeper insights into their behavior and interactions. 

\begin{figure}
    \vspace{1.5mm}
    \centering
    \includegraphics[width=\columnwidth]{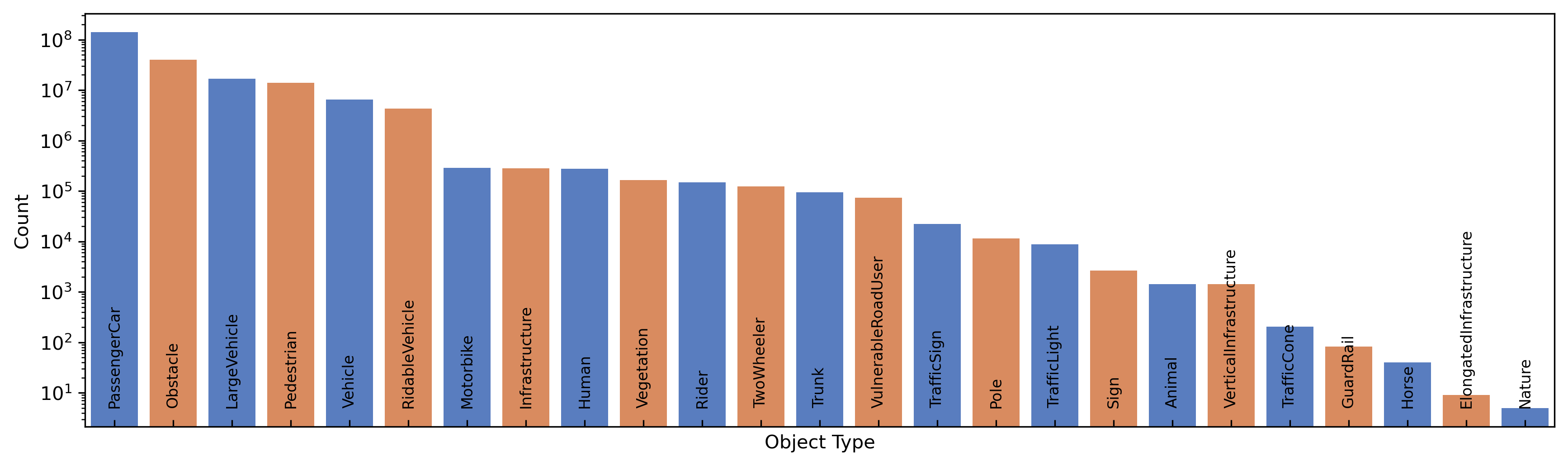}
    \caption[Distribution of Object Types in Raw Data]{Distribution of object types in raw data.}    \label{fig:raw_type_dist}
\end{figure}

\begin{figure}[ht]
    \centering
    \includegraphics[width=0.85\columnwidth]{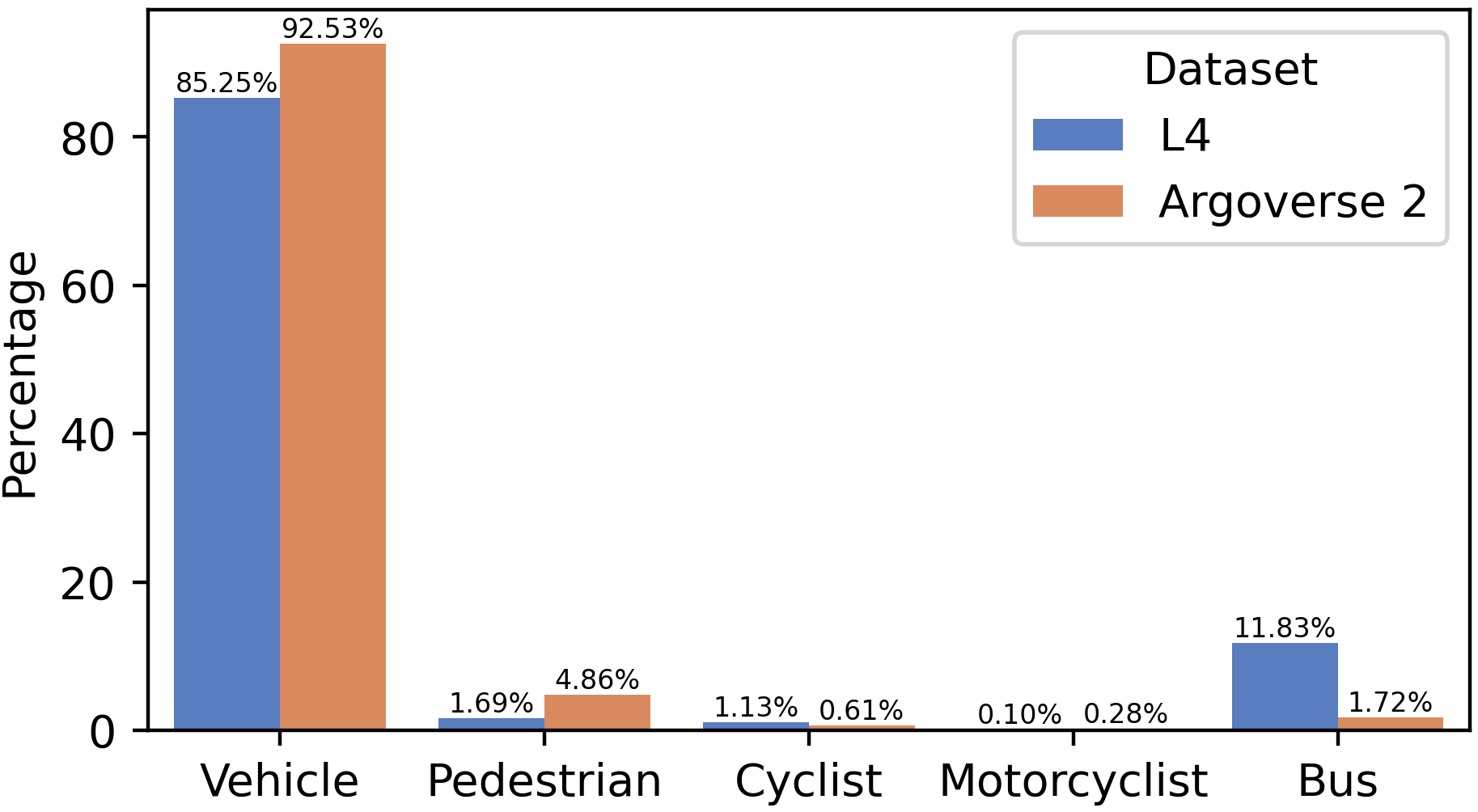}
    %\captionsetup{width=0.93\linewidth}
    \caption[Comparison of Track Types Across Argoverse 2 and L4]{Comparison of the type distribution across scored tracks and focal tracks in the L4 and Argoverse~2 dataset.}
    \label{fig:data_compare_scored_focal_type}
\end{figure}

The recorded objects are classified into 25 different hierarchical object classes. The detailed distribution can be found in \cref{fig:raw_type_dist}. However, the object hierarchy is not important for our use case because the 25 classes are later sorted into the five object classes of the AV2 dataset. The object type distribution of the final L4 dataset can be seen in \cref{fig:data_compare_scored_focal_type}. Similar to AV2 the vehicle class is by far the most common. The L4 dataset has a lot more bus objects. This is because in our raw data buses and trucks are both classified as large vehicles. This makes sense since they have similar movement patterns and have to follow similar traffic rules which are only applicable to large vehicles.

The dataset is enriched with various \textbf{Agent Features} not present in  AV2, which include:
\begin{itemize}
    \item \textbf{Frenet Frame Attributes}: These attributes provide a detailed perspective on the position and motion of each object relative to the lane geometry, including lateral and longitudinal velocities, accelerations, and their variances.
    \item \textbf{Distance to Objects Attributes}: Proximity metrics are included to measure distances to critical road features, such as crosswalks, stop signs, and traffic lights.
    \item \textbf{Additional Geometric Attributes}: This includes measurements such as the width and length of bounding boxes, yaw rates, and variances in position and velocity.
\end{itemize}

In addition to the agent features, the HD map incorporates several additional \textbf{Map Features} not included in AV2:
\begin{itemize}
    \item \textbf{Extra Lane Segments}: Additional lane segments, such as shoulder and parking lanes.
    \item \textbf{Maximum Speed Limit}: Each lane segment is annotated with maximum speed limits.
    \item \textbf{Stop Lines}: Real and virtual stop lines are included. Virtual stop lines indicate implicit traffic rules such as yielding to oncoming traffic on a left turn.
\end{itemize}

\subsection{Dataset Creation}

To ensure compatibility with AV2, we structure the L4 dataset to align with the AV2 format. The necessary processing steps are detailed in the following sections. Raw data undergoes quality control to ensure only valid data is included. Raw continuous recording are sorted into train, validation and test splits. These continuous recordings are then segmented into standardized 11-second scenarios with HD map context. 

\subsection*{1) Quality Control and Splitting}

Before converting the raw recordings, we perform a series of sanity checks to verify the integrity of the data. These include the detection and removal of recordings with:
\begin{itemize}
    \item Corrupt or missing sensor data
    \item Abnormally sampled sequences (e.g., 20 Hz or duplicated timestamps)
    \item Scenarios with excessive or inconsistent agent IDs due to sensor malfunction or tracking errors
\end{itemize}

\begin{figure}
    \vspace{1.5mm}
    \centering
    \subfloat[]{\includegraphics[width=0.49\columnwidth]{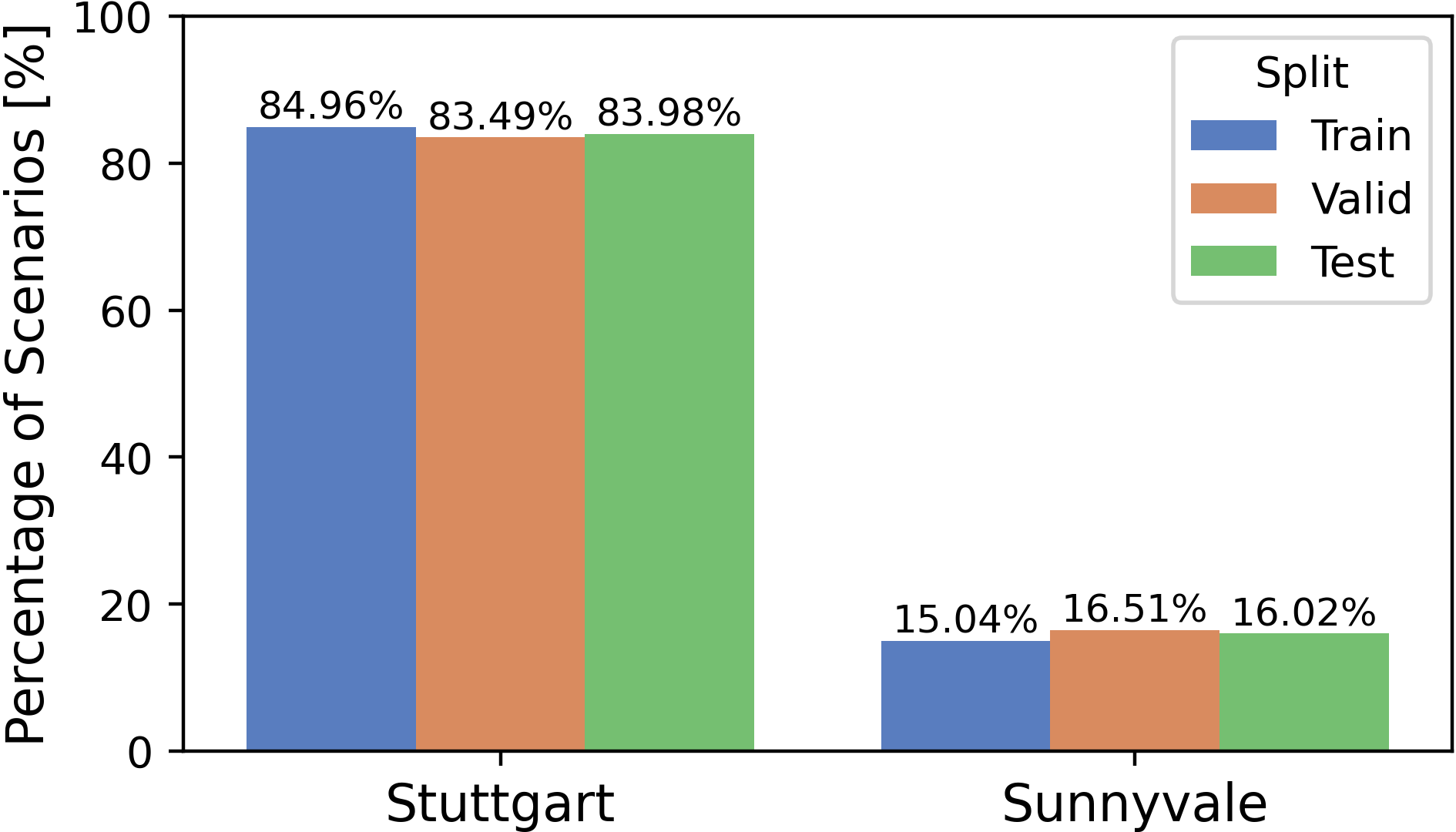}\label{fig:datasplit_city_env_city}}
    \hfill
    \subfloat[]{\includegraphics[width=0.49\columnwidth]{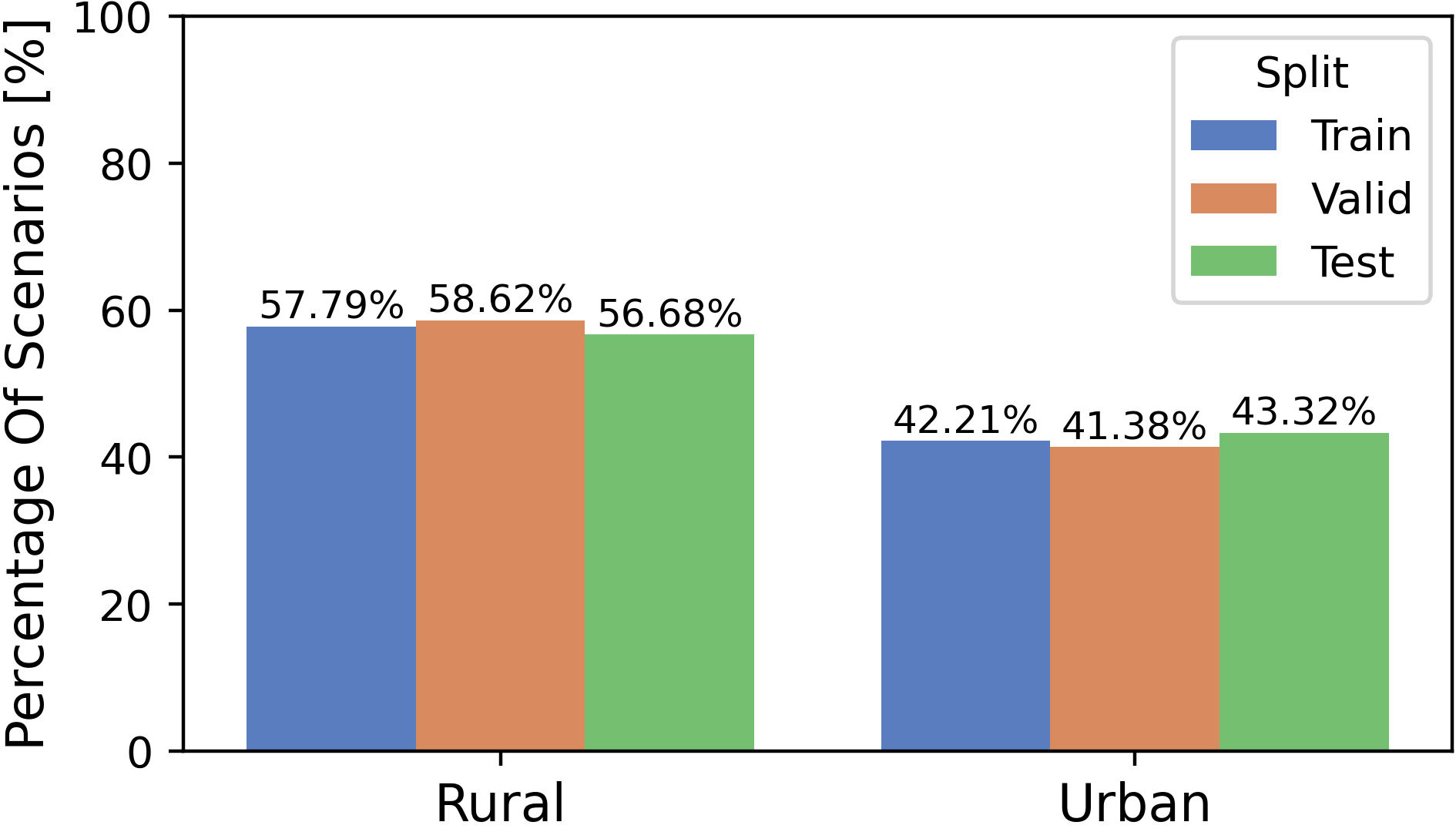}\label{fig:datasplit_city_env_city_env}}
    \caption[Dataset Splits: City and Scenario Environment Distributions]{Exact distributions in dataset splits for a) regions and b) environment.}
    \label{fig:datasplit_city_env}
\end{figure}

Following this, we split the dataset into three sets: \textit{training}~(80\%), \textit{validation} (10\%), and \textit{test} (10\%). To ensure temporal separation between the splits we do not split continuous recording files. We made sure that the splits are balanced across:
\begin{itemize}
    \item \textbf{Geographic location}: Germany vs. US (see \cref{fig:datasplit_city_env_city})
    \item \textbf{Environment}: Urban vs. Rural (see \cref{fig:datasplit_city_env_city_env})
    \item \textbf{Object type diversity}
\end{itemize}

% \begin{figure}[ht]
%     \centering
%     \includegraphics[width=\textwidth]{figures/4_methodology/environment_city_distribution.png}
%     \caption[Dataset Splits: City and Scenario Environment Distributions]{On the left, the distribution of cities, and on the right, the distribution of scenario environments across the dataset splits.}
%     \label{fig:datasplit_city_env}
% \end{figure}

\subsection*{2) Scenario Segmentation and Formatting}

Each recording file is processed to extract \textit{11-second scenarios}, consistent with the AV2 standard. Temporal discontinuities exceeding 150 ms are used to segment the continuous driving sequences. Only scenarios with a complete \textit{focal track} are retained.

Tracks within each scenario are categorized into:
\begin{itemize}
    \item \textbf{Focal Track:} Primary prediction target for single-agent predictions, prioritized for object type diversity.
    \item \textbf{Scored Tracks:} Dynamic agents interacting with the focal object. Used in multi-agent predictions.
    \item \textbf{Unscored Tracks:} Static objects and ego vehicle.
    \item \textbf{Track Fragments:} Objects which are only partially observed over the entire 11s in a scenario.
\end{itemize}

\subsection*{3) HD Map Extraction}

\begin{figure}
    \vspace{1.5mm}
    \centering
    \includegraphics[width=\columnwidth]{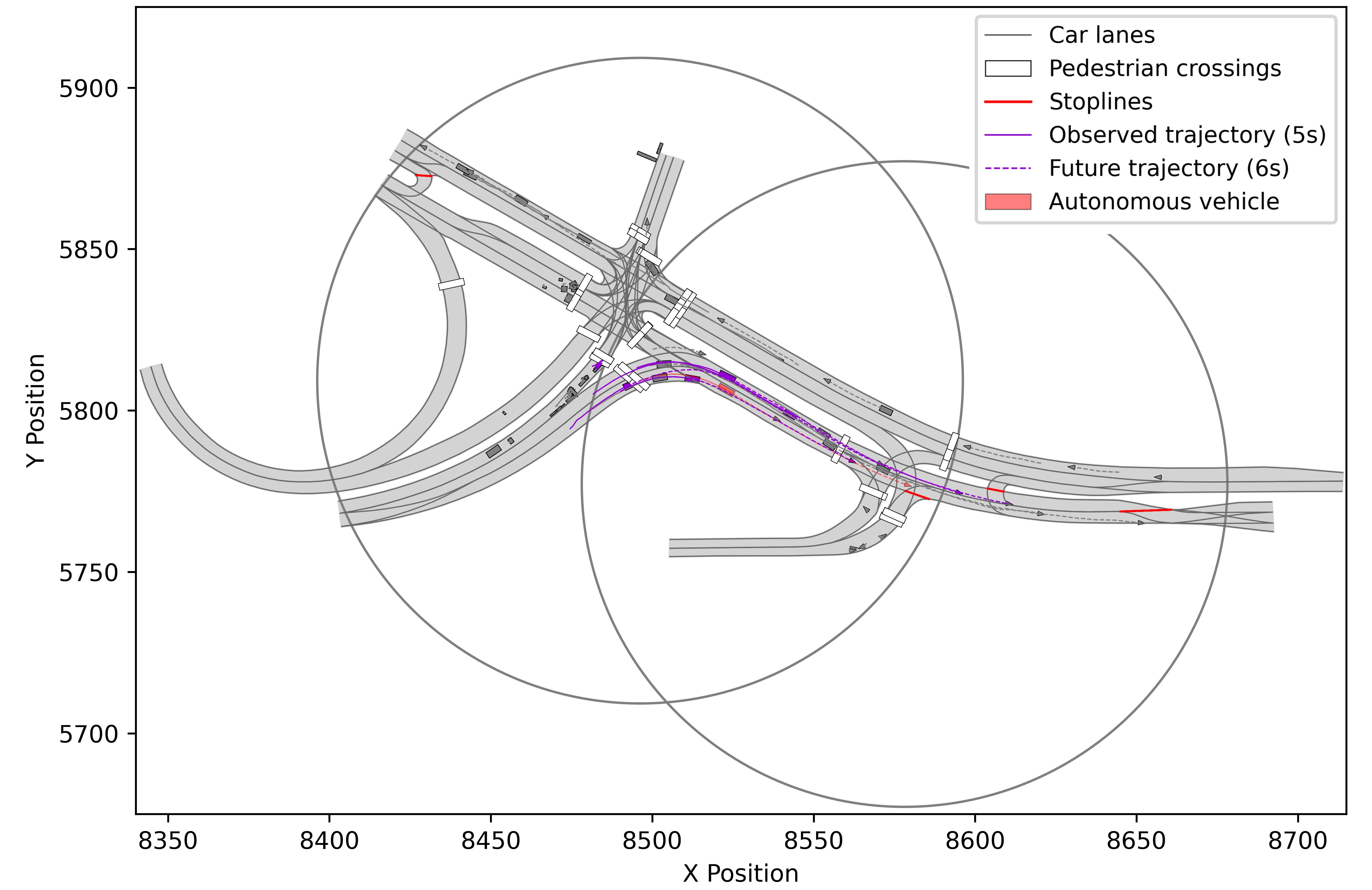}
    \caption[Illustration of a Driving Scenario]{Illustration of a driving scenario with highlighted lanes, pedestrian crossings, and stop lines. Purple lines represent the trajectories of the scored and focal tracks with their respective bounding boxes, while gray lines indicate other track fragments. The large circles depict 100m radiuses at the first and last positions of the automated vehicle.}
    \label{fig:scenario_example}
\end{figure}

For each scenario, we generate HD map context data by extracting all relevant elements within a \textit{100-meter radius} around the ego vehicle's trajectory (see \cref{fig:scenario_example}). This is exactly the same HD map size as in AV2. Our HD map includes:
\begin{itemize}
    \item Lane segments and lane boundaries
    \item Pedestrian crossings and stop lines %(converted into polylines)
    \item Extra map features such as maximum speed limits, parking/shoulder lanes, and stop/yield annotations
\end{itemize}

\subsection*{4) Dataset Analysis}

\begin{figure}
    \centering
    \includegraphics[width=0.8\columnwidth]{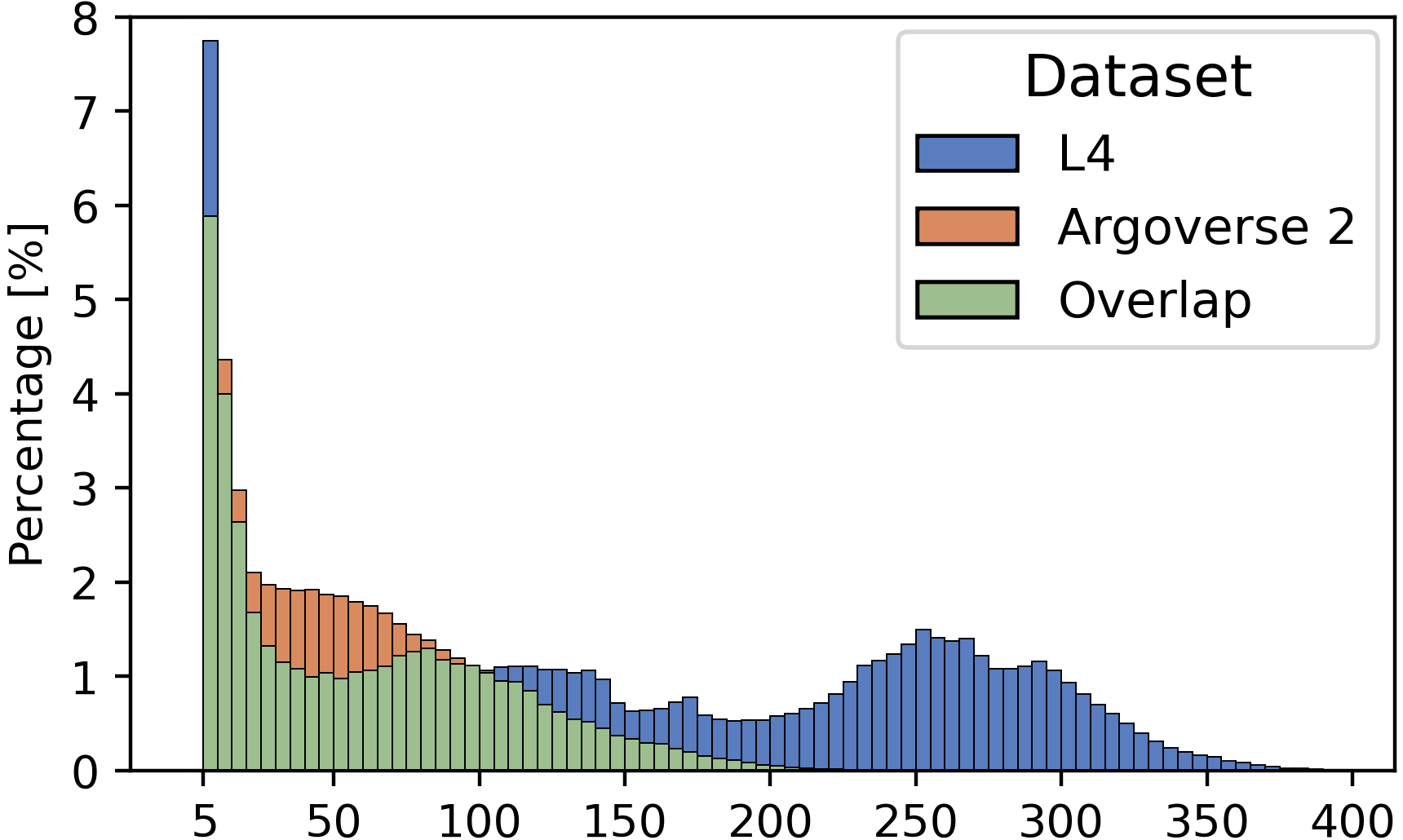}
    \caption{Trajectory length (m) for scored and focal tracks with 11s duration. Stationary objects under 5m trajectory length are excluded.}
    \label{fig:data_compare_distance}
    \vspace{-6mm}
\end{figure}

When comparing the distribution of trajectory lengths of our L4 dataset with the AV2 dataset we can see that our dataset spans a much wider range of trajectory lengths for the 11s long scenarios, as can be seen in \cref{fig:data_compare_distance}. This directly translates to a wider range of speed profiles of which the L4 dataset contains a balanced amount corresponding to different driving situations ranging from slow speeds in urban areas to high speeds in rural areas containing country roads and highways. The longest trajectory in the L4 dataset is approximately 470m long which corresponds to a speed of 150km/h. The AV2 scenarios prioritize city centers, emphasizing urban driving conditions. Here the top speed is much lower with approximately 70km/h. 

\begin{figure}
    \vspace{1.5mm}
    \centering
    \includegraphics[width=0.75\columnwidth]{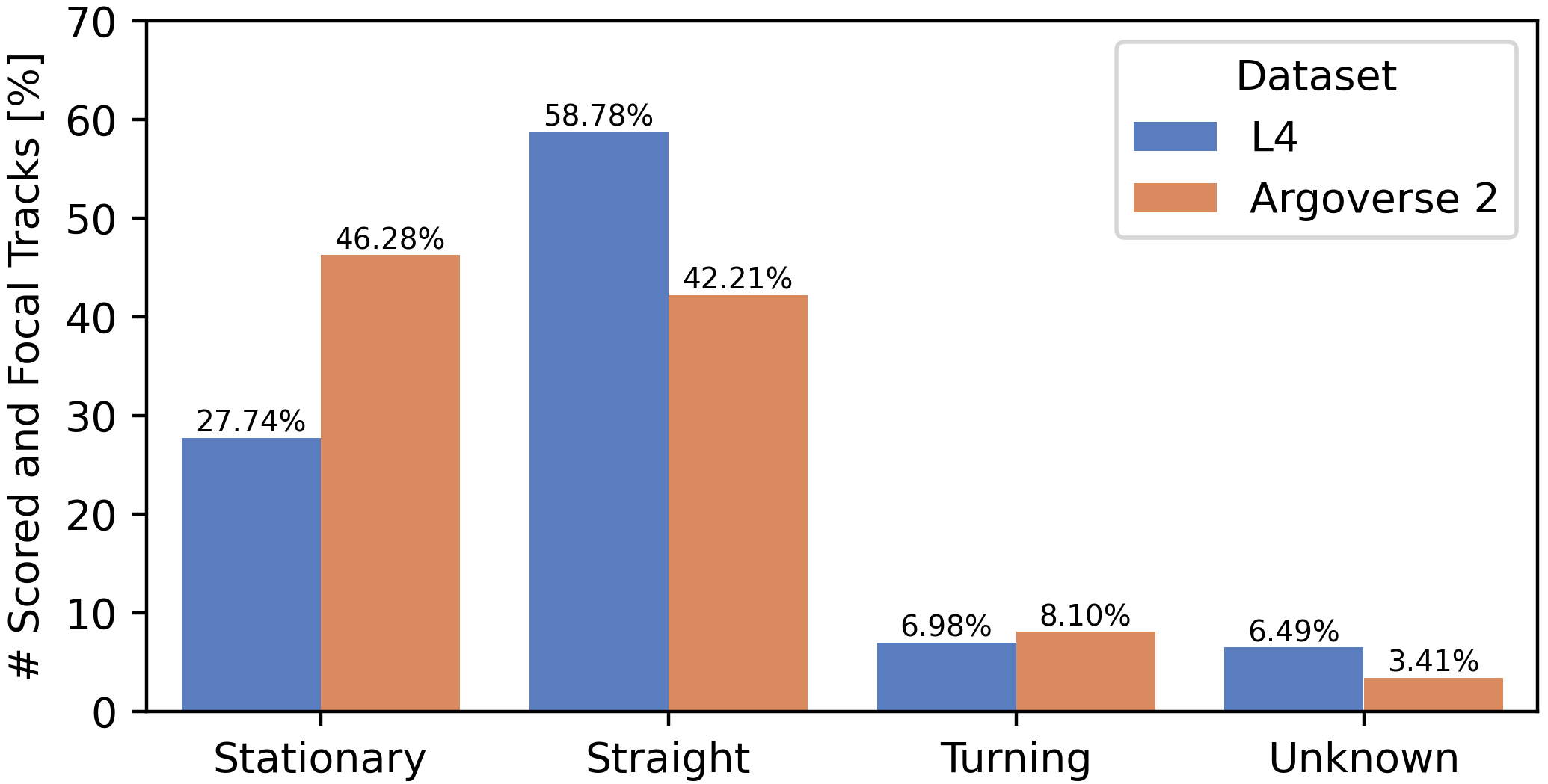}
    %\captionsetup{width=0.93\linewidth}
    \caption{Distribution of maneuvers in L4 and AV2.}
    \label{fig:data_compare_manuever}
\end{figure}

\cref{fig:data_compare_manuever} shows the distribution of different maneuvers in the datasets. Here we observe again the bigger focus on urban driving in AV2. AV2 contains many more stationary actors which are for example waiting at a red light or are parked. The classification of these maneuvers was based on analyzing the ratio between the Euclidean distance from the trajectory's start to end points and the total trajectory length, as well as identifying short trajectories ($\leq$ 5m) as stationary.

\begin{figure}
    \centering
    \subfloat[L4]{\includegraphics[width=0.49\columnwidth]{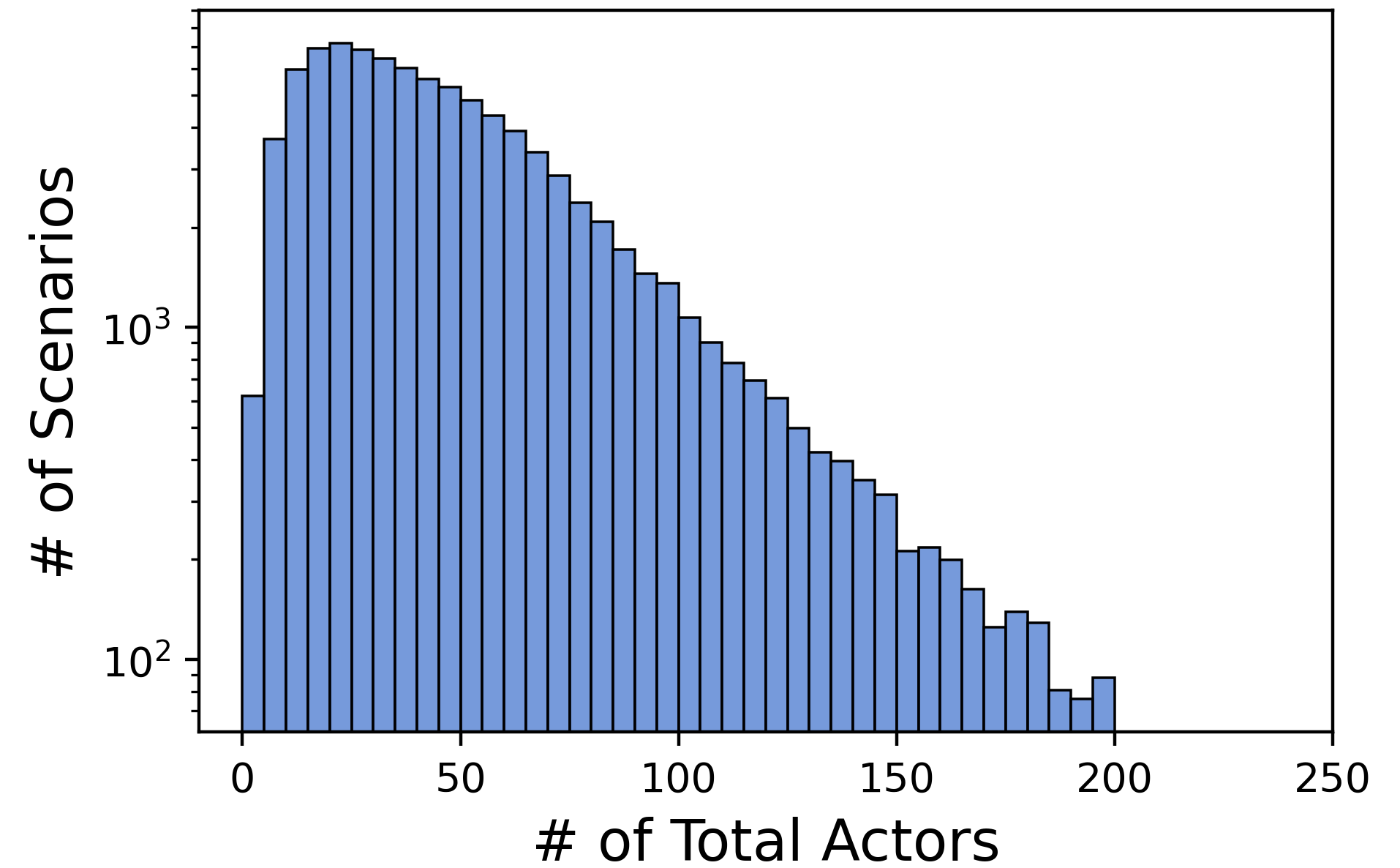}\label{fig:data_compare_track_dist_L4}}
    \hfill  % Space between images
    \subfloat[AV2 (taken from \cite{Argoverse2_Additional_Info})]{\includegraphics[width=0.49\columnwidth]{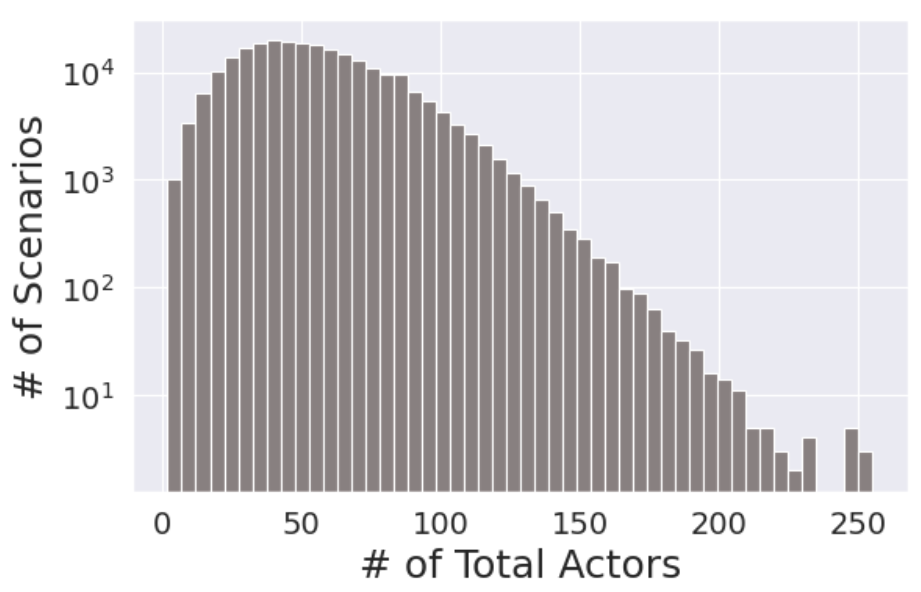}\label{fig:data_compare_track_dist_Argo}}
    \caption[Comparison of Actors Distribution Across Argoverse 2 and L4]{Number of actors per scenario.}
    \label{fig:data_compare_track_dist}
    %\vspace{-1mm}
\end{figure}
\begin{figure}[h]
    \vspace{1.5mm}
    \centering
    \subfloat[L4]{\includegraphics[width=0.49\columnwidth]{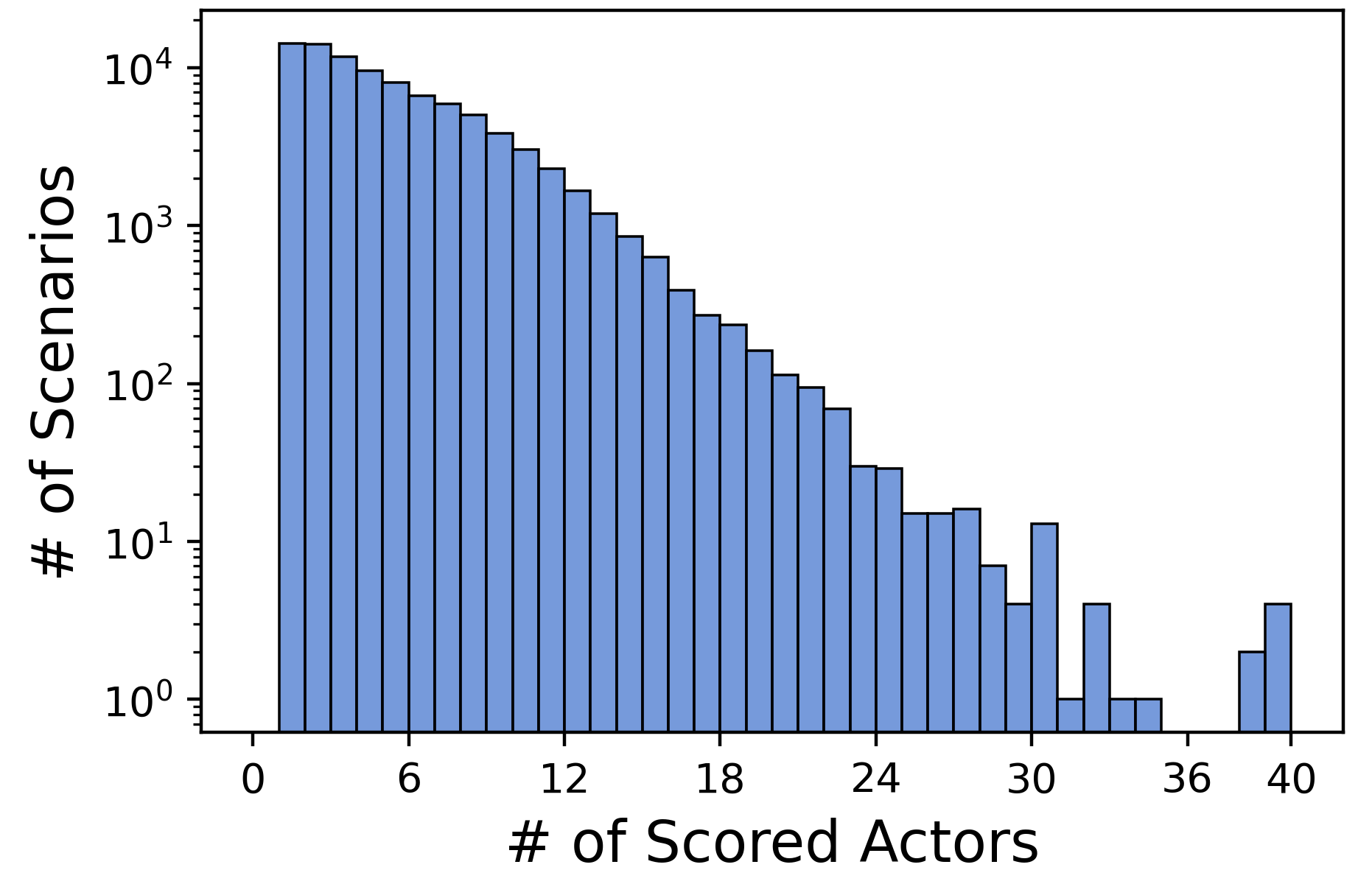}\label{fig:scored_track_dist_L4}}
    \hfill  
    \subfloat[AV2 (taken from \cite{Argoverse2_Additional_Info})]{\includegraphics[width=0.49\columnwidth]{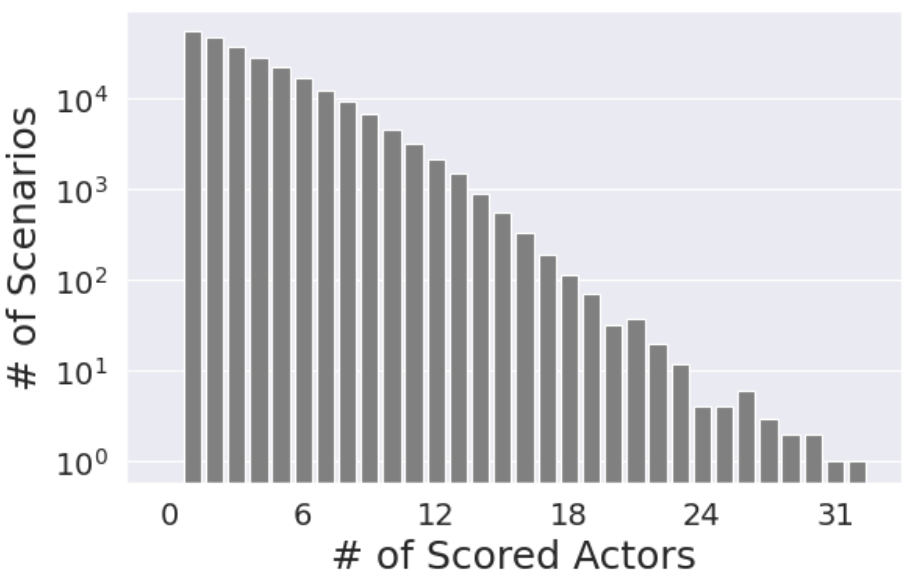}\label{fig:scored_track_dist_Argo}}
    \caption[Comparison of Scored Tracks Across Argoverse 2 and L4]{Number of scored and focal tracks per scenario.}
    \label{fig:data_compare_scored_dist}
\end{figure}

When looking at the number of actors per scenario (see \cref{fig:data_compare_track_dist}) and the number of scored and focal tracks per scenario (see \cref{fig:data_compare_scored_dist}) we can observe similar patterns in both datasets. Since AV2 is a larger dataset the absolute number of scenarios is also larger than in the L4 dataset. When looking at the distribution of all observed actors, even actors only observed for a few timesteps, we can see that AV2 tends to have more actors in its scenarios. But when only looking at the scored and focal tracks both datasets have a more similar distribution. However, the L4 dataset has a bunch of scenarios with up to 40 scored actors which are not present in the AV2 dataset.

\begin{figure}
    %\vspace{1.5mm}
    \centering
    \includegraphics[width=0.7\columnwidth]{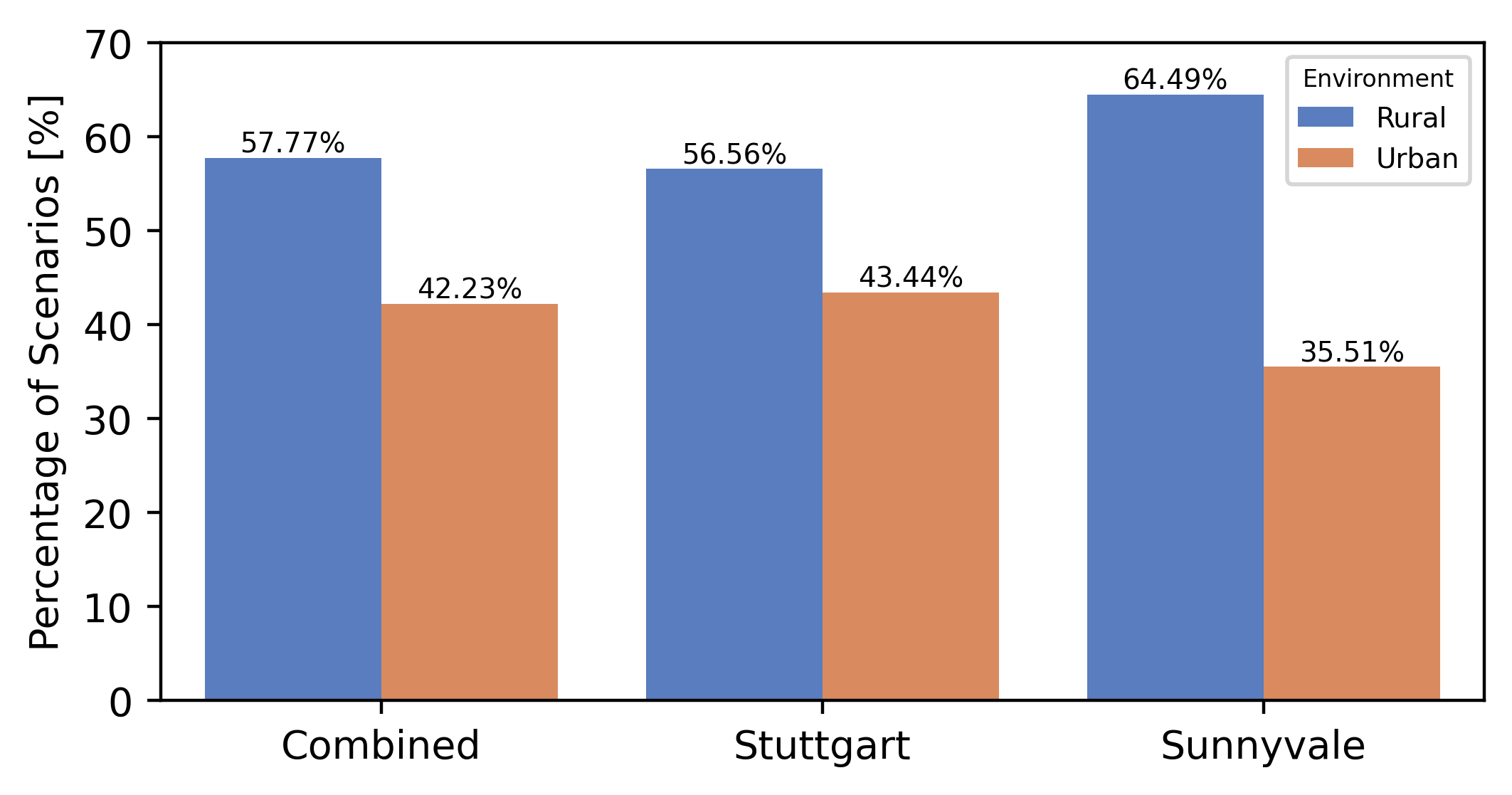}
    \caption[Comparison of Scenario Environments Across L4]{Comparison of the environments in our dataset. 
    }
    \label{fig:scenario_environment}
    \vspace{-2mm}
\end{figure}

Lastly, if we compare the Stuttgart fraction of the L4 dataset with the Sunnyvale fraction, we can see that the recordings in Stuttgart are more balanced between rural and urban driving (see \cref{fig:scenario_environment}). In Sunnyvale, more data was collected on the highway as can be seen in the recording areas in \cref{fig:unique_roadways_sunnyvale}.

%\vspace{\baselineskip}

\section{Evaluation}
For our evaluation, we use a multi-agent setting. Predictions are generated for all focal and scored agents in the dataset. We use the standard single-agent metrics used in the AV2 benchmark \cite{Argoverse2_Original, Argoverse2Challenge}. These are calculated for each agent and averaged across the entire dataset. 

Let $\mathcal{S}$ denote the set of scenes, $\mathcal{A}(s)$ the set of agents in scene $s \in \mathcal{S}$, 
and $M_a$ the metric value for agent $a$. The final metric $M$ is then:

\begin{equation}
    M = \frac{\sum_{s \in \mathcal{S}} \sum_{a \in \mathcal{A}(s)} M_a}{\sum_{s \in \mathcal{S}} \lvert \mathcal{A}(s) \rvert} .
\end{equation}

\subsection{Additional Features}\label{sec:metrics}
As described in \cref{sec:data}, our dataset contains more details than other datasets. We test the impact of these additional features by training and evaluating QCNet using multiple subsets of these features:

\begin{itemize}
    \item \textbf{Baseline:} No additional features. Only features present in AV2 are used.
    \item \textbf{Enhanced Lanes:} Contains additional lane segments with maximum speed limits to provide a richer contextual environment.

    \item \textbf{Enhanced Stop Lines:} Integrates additional stop/yield line data to help refine vehicle dynamics at intersections. 
    
    \item \textbf{Enhanced Agents:} Extends the agent-specific data by introducing more granular geometric details.
    
    \item \textbf{Full Integration:} A comprehensive mix of all available features to assess their collective impact.
    
    \item \textbf{Focused Integration:} A strategic selection that combines Enhanced Stop Lines, Enhanced Lanes, and selective agent features. This approach is designed to reduce redundancy by focusing only on additional agent features that cannot be obtained based on other attributes. In particular, Frenet frame attributes can be derived from agent positions and corresponding HD map data. The same applies to distances to objects, as these can be calculated from spatial relationships between agents and objects.
\end{itemize}

The results of the different feature sets are documented in \cref{tab:results_training_l4_features}. We were not able to observe any notable changes in any metric for any of the tested sets of additional features. Even when looking at individual scenarios in which these additional features could be useful we do not observe notable improvement. One of these scenarios is analyzed in \cref{sec:qualitative} (\cref{fig:quali_feat_base}). This leads us to the following conclusion: Either, the network is already powerful enough to infer the additional features from the limited baseline features or it is not capable of leveraging the additional features. Either way, the additional features provide no noteworthy benefit. This validates the chosen feature sets of other public datasets. 

\begin{table*}
\vspace{1.5mm}
\captionsetup{width=0.8\linewidth}
    \caption[Training Results for Additional Features]{Performance evaluation of QCNet trained on various feature sets.}
\centering
\begin{tabular}{@{}l|cccc|ccc@{}}
    \toprule
    Feature Set & $\text{b-minFDE}$$\downarrow$ & $\text{minADE}_6$$\downarrow$ & $\text{minFDE}_6$$\downarrow$ & $\text{MR}_6$$\downarrow$ & $\text{minADE}_1$$\downarrow$ & $\text{minFDE}_1$$\downarrow$ & $\text{MR}_1$$\downarrow$ \\
    \midrule
    Baseline            & 1.29 & 0.47 & 0.69 & 0.07 & 0.99 & 2.39 & 0.35\\
    \midrule
    Enhanced Lanes      & 1.30 & 0.47 & 0.69 & 0.07 & 1.00 & 2.41 & 0.35\\
    Enhanced Stoplines  & 1.30 & 0.47 & 0.69 & 0.07 & 1.01 & 2.42 & 0.35\\
    Enhanced Agent      & 1.29 & 0.47 & 0.68 & 0.07 & 0.99 & 2.37 & 0.35\\
    \midrule
    Full Integration    & 1.30 & 0.47 & 0.69 & 0.07 & 0.99 & 2.37 & 0.35\\
    \midrule
    Focused Integration & 1.29 & 0.47 & 0.68 & 0.07 & 0.99 & 2.38 & 0.35\\
    \bottomrule
\end{tabular}
    
    \label{tab:results_training_l4_features}
\end{table*}

\subsection{Transfer Between Datasets}

We were also interested in how the knowledge of a network trained on one dataset transfers to another dataset. To evaluate this transfer we compare three QCNet models. One is trained on the baseline L4 dataset only using the features available in the AV2 dataset. Another one is trained on the AV2 dataset. Lastly, we take the model trained on the AV2 set and continue training it on our L4 dataset. 

Although the dataset structures are identical, there are still major differences. The biggest is the location. AV2 only uses US data, while our L4 dataset has a focus on Germany with a smaller fraction of US data. The L4 dataset also includes rural scenarios while the AV2 dataset only focuses on urban driving. Other differences can be found in the HD map. For example how the lanes are divided into lane segments. 

The experimental outcomes of this domain shift analysis are presented in \cref{tab:results_transferlearning}. Consistent with expectations, models trained exclusively on a single dataset achieve optimal performance on their corresponding evaluation dataset. Notably, predictive accuracy on the L4 dataset is generally lower than on AV2, suggesting that the L4 dataset poses greater inherent challenges. This hypothesis is supported by two key factors: 

(1) the L4 dataset encompasses a broader spectrum of speed profiles, ranging up to 150km/h, introducing greater variability in motion patterns that complicate trajectory prediction, especially since endpoint predictions have a strong impact on the evaluation metrics used. 

(2) complex traffic scenarios, like roundabouts, prevalent in L4 but underrepresented in AV2, present additional challenges due to their dynamic navigation requirements, which are less frequently encountered in AV2’s training data. This is also shown in our qualitative analysis in \cref{sec:qualitative}.

Further reinforcing this conclusion, the performance degradation of the L4-trained model evaluated on AV2 is less severe than the reverse scenario (AV2-trained model evaluated on L4), highlighting L4’s broader operational complexity. Additionally, the model pre-trained on AV2 and fine-tuned on L4 exhibits only marginally inferior performance on L4 compared to the L4-only model, while retaining substantial knowledge from AV2. This dual-training approach achieves significantly better results on AV2 than the L4-only model, demonstrating that L4’s diversity imposes stricter learning requirements but enables more robust generalization when combined with AV2’s data.

\begin{table*}
\vspace{1.5mm}
\centering
\caption[Training Results Across Different Datasets]{Performance evaluation of QCNet models trained and validated on different datasets.}
\begin{tabular}{@{}l||cc|cc|cc|cc||cc|cc|cc@{}}
    \toprule
    Training & 
    \multicolumn{2}{c|}{$\text{b-minFDE}$$\downarrow$} & 
    \multicolumn{2}{c|}{$\text{minADE}_6$$\downarrow$} & 
    \multicolumn{2}{c|}{$\text{minFDE}_6$$\downarrow$} &
    \multicolumn{2}{c||}{$\text{MR}_6$$\downarrow$} & 
    \multicolumn{2}{c|}{$\text{minADE}_1$$\downarrow$} & 
    \multicolumn{2}{c|}{$\text{minFDE}_1$$\downarrow$} & 
    \multicolumn{2}{c}{$\text{MR}_1$$\downarrow$} \\
    Dataset & L4 & AV2 
    & L4 & AV2 
    & L4 & AV2 
    & L4 & AV2 
    & L4 & AV2 
    & L4 & AV2 
    & L4 & AV2\\
    \midrule
    L4         & \textbf{1.29} & 1.54 & \textbf{0.47} & 0.59 & \textbf{0.69} & 0.99 & \textbf{0.07} & 0.13 & 0.99 & 1.49 & 2.39 & 3.82 & \textbf{0.35} & 0.38  \\

    AV2 \& L4  & 1.31 & 1.43 & 0.48 & 0.51 & 0.70 & 0.82 & \textbf{0.07} & 0.11 
    & \textbf{0.98} & 1.26 & \textbf{2.35} & 3.17 & \textbf{0.35} & 0.35\\
    AV2      & 1.99 & \textbf{1.22} & 1.04 & \textbf{0.37} & 1.37 & \textbf{0.60} & 0.20 & \textbf{0.07} & 2.15 & \textbf{0.83} & 5.21 & \textbf{2.08} & 0.52 & \textbf{0.28}\\
    \bottomrule
\end{tabular}
    
    \label{tab:results_transferlearning}
\end{table*}

\subsection{Transfer Between Locations}

For a more direct comparison between different locations, we split our dataset based on countries. This eliminates the impact of the domain shift affecting the results of the previous section. A downside of this approach is the imbalance in our L4 dataset regarding location distribution. The dataset contains more data for the Stuttgart area than for Sunnyvale. 

\cref{tab:results_training_geographic_subsets_l4} shows the results for the whole L4 dataset and the locations subsets for Stuttgart and Sunnyvale. Using the entire dataset for training yields the best results across the board. Interestingly, the model trained on the Stuttgart set comes in second place for most metrics, even when evaluated on the Sunnyvale test set. The model trained on the Sunnyvale set performs significantly worse on the Stuttgart test set and on the whole L4 set. Even on the Sunnyvale test set the performance is worse than the model trained on the Stuttgart set. 

We draw the following conclusions from these results: 

The Sunnyvale dataset lacks the size and diversity to train a model like QCNet to its limits. The Stuttgart set allows the model to generalize to the point that it outperforms the model solely trained on Sunnyvale data in its own domain. 

This also indicates that a model trained only on data from one country can generalize enough to perform adequately in another country.

However, the model trained on the whole dataset performs the best in all domains. This indicates that a more diverse dataset spanning multiple countries can benefit not only the overall performance but also in specific domains. By including the smaller Sunnyvale dataset the performance increases for the Stuttgart test set too. To gain these benefits a smaller dataset from an additional country is sufficient.

\begin{table*}
\centering
\captionsetup{width=0.8\linewidth}
\caption{Performance evaluation of QCNet trained and evaluated on different geographic subsets in the L4 dataset. Each model is trained with all additional features.}
\begin{tabular}{@{}l|ccc|ccc|ccc|ccc@{}}
    \toprule
    Training Set &
    \multicolumn{3}{c|}{$\text{b-minFDE}$$\downarrow$} & 
    \multicolumn{3}{c|}{$\text{minADE}_6$$\downarrow$} & 
    \multicolumn{3}{c|}{$\text{minFDE}_6$$\downarrow$} &
    \multicolumn{3}{c}{$\text{MR}_6$$\downarrow$}\\ 
    & L4 & STG & SVL
    & L4 & STG & SVL 
    & L4 & STG & SVL
    & L4 & STG & SVL \\
    \midrule
    L4   
    & \textbf{1.30} & \textbf{1.32} & \textbf{1.17}
    & \textbf{0.47} & \textbf{0.49} & \textbf{0.35} 
    & \textbf{0.69 }& \textbf{0.71} & \textbf{0.58} 
    & \textbf{0.07} & \textbf{0.07} & \textbf{0.07}\\
    Stuttgart (STG)
    & 1.33 & 1.34 & 1.30
    & 0.48 & 0.50 & 0.39
    & 0.71 & \textbf{0.71} & 0.71
    & \textbf{0.07} & \textbf{0.07} & 0.08\\
    Sunnyvale (SVL) 
    & 2.25 & 2.42 & 1.35
    & 0.85 & 0.93 & 0.43
    & 1.61 & 1.78 & 0.71
    & 0.22 & 0.24 & 0.09\\
    \bottomrule
\end{tabular}
    
    \label{tab:results_training_geographic_subsets_l4}
\end{table*}

\subsection{Qualitative Results}\label{sec:qualitative}

\begin{figure}
    \centering
    \subfloat[Baseline (no additional features)]{%
        \includegraphics[width=\columnwidth]{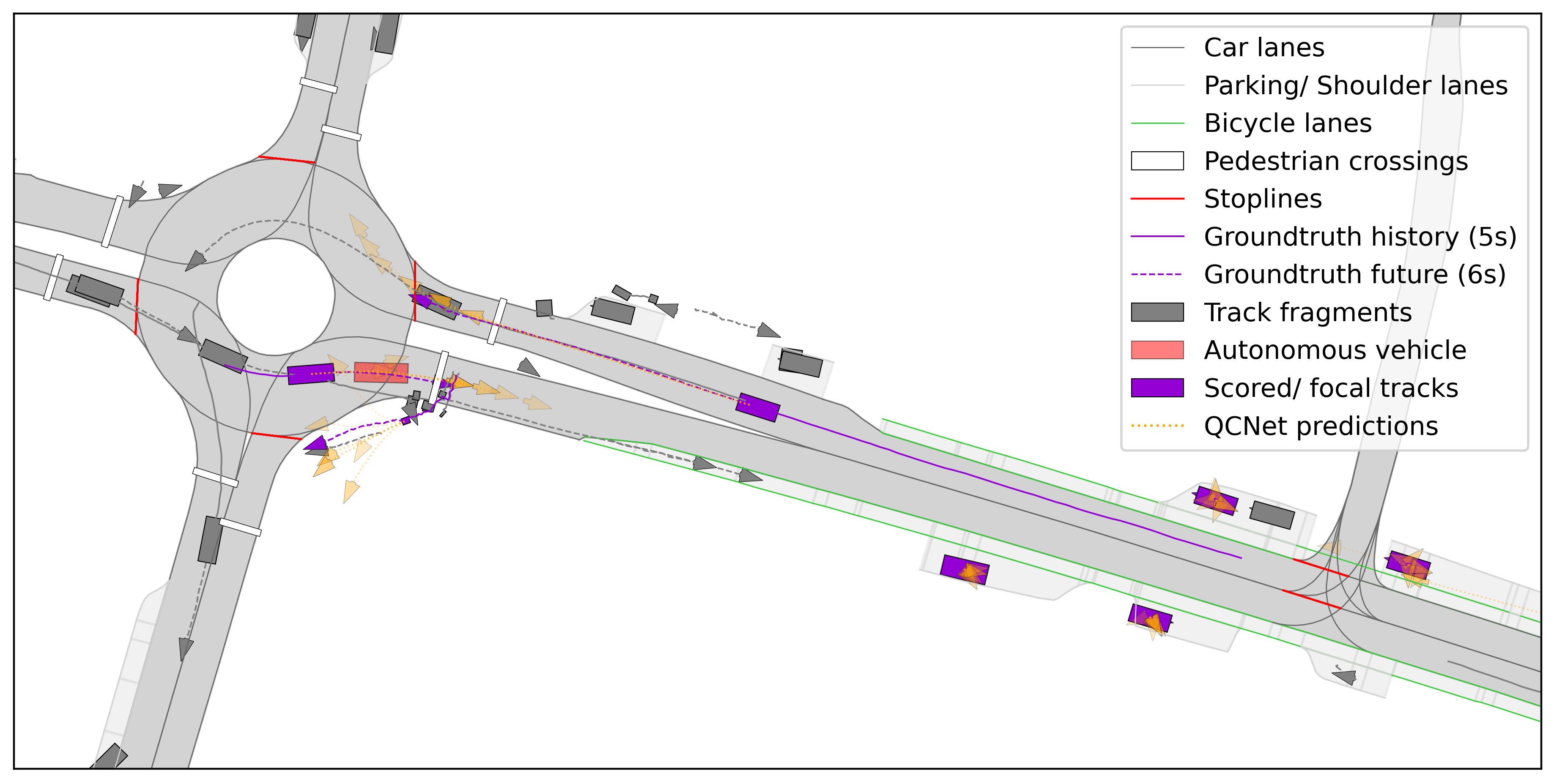}
        \label{fig:quali_feat_base_baseline}
    }\\ 
    \subfloat[Full integration (all additional features)]{%
        \includegraphics[width=\columnwidth]{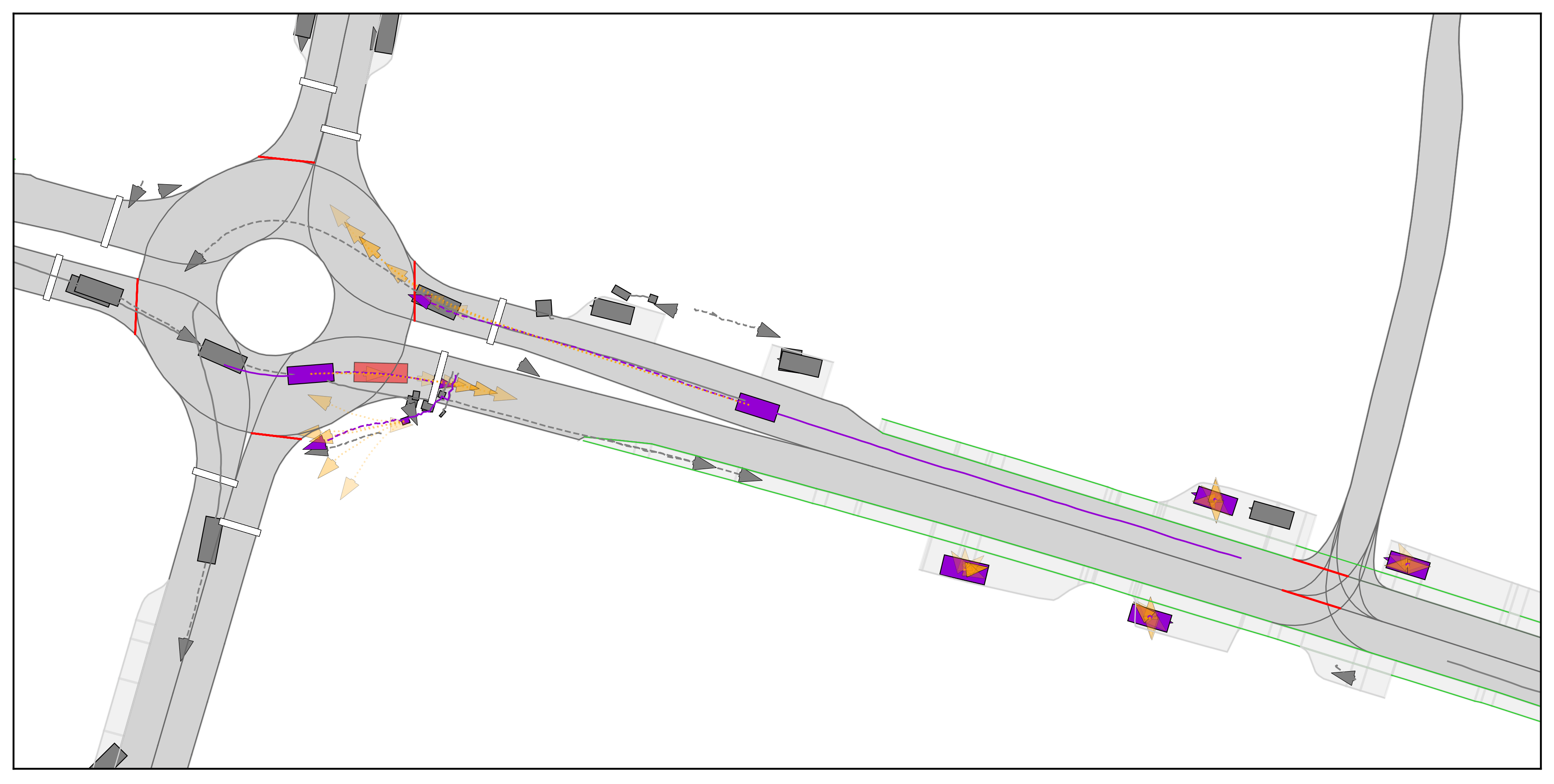}
        \label{fig:quali_feat_base_all_features}
    }
    \caption{Comparison between models trained on different L4 feature sets in a scenario of the L4 test set.}
    \label{fig:quali_feat_base}
\end{figure}

Lastly, we look into some qualitative results. \cref{fig:quali_feat_base} shows a comparison between two models trained using different feature sets. The scenario shows a roundabout that has additional stop lines at its entries. The ground truth trajectory for the car in the center of the image driving towards the roundabout ends before entering the roundabout. The predictions of the baseline model seen in \cref{fig:quali_feat_base_baseline} has a stronger tendency to stop there despite not using the stop line information whereas the model using all additional features (including the stop line features) has a stronger tendency to enter the roundabout. Overall, this is not a major difference despite this being a situation in which we would expect the additional features to be beneficial. This aligns with our findings in \cref{sec:metrics}. 

\begin{figure}
    \centering
    \subfloat[L4]{%
        \includegraphics[width=0.47\columnwidth]{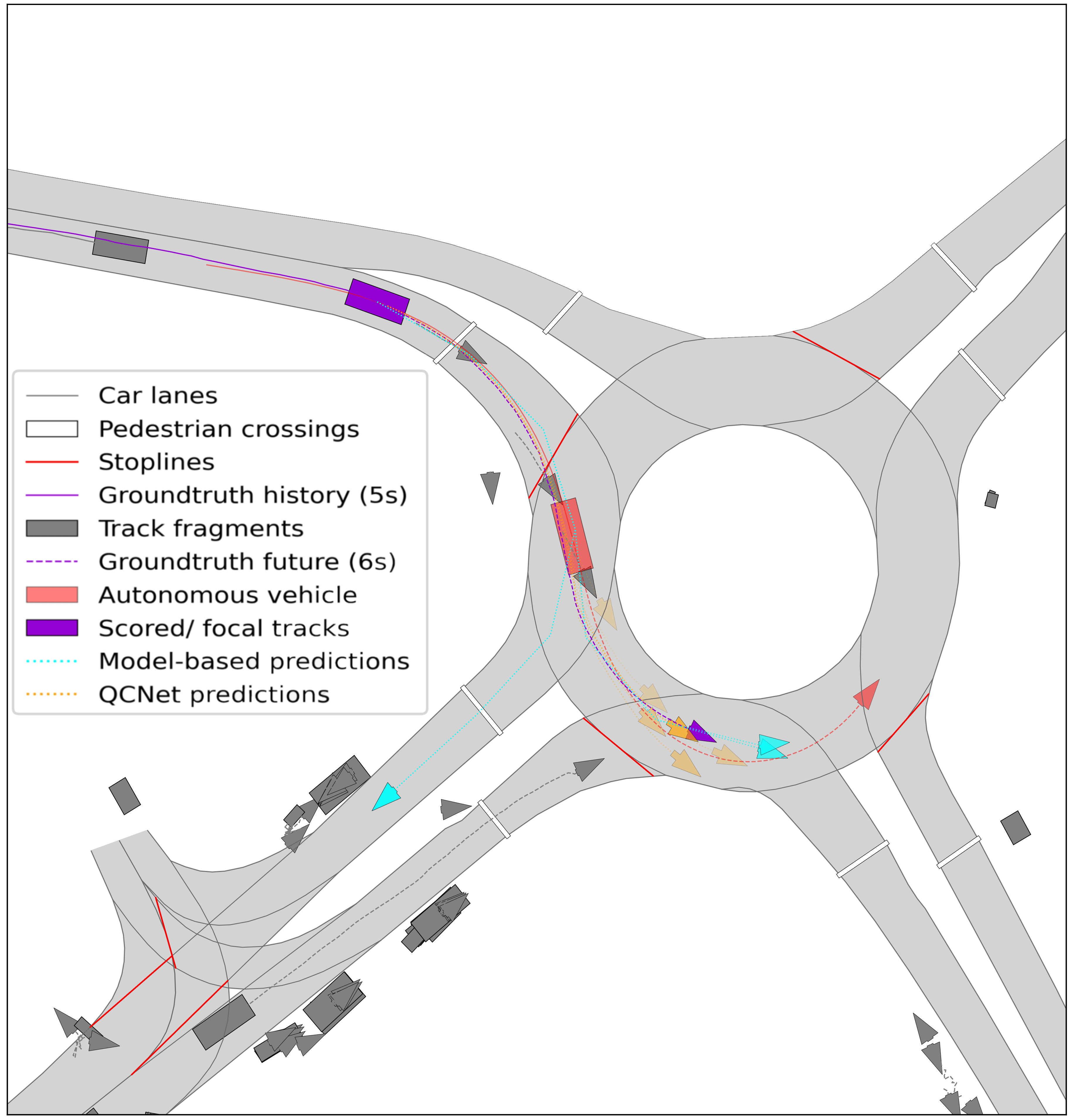}
        \label{fig:quali_argo_base_model_l4}
    }\hfill 
    \subfloat[AV2]{%
        \includegraphics[width=0.47\columnwidth]{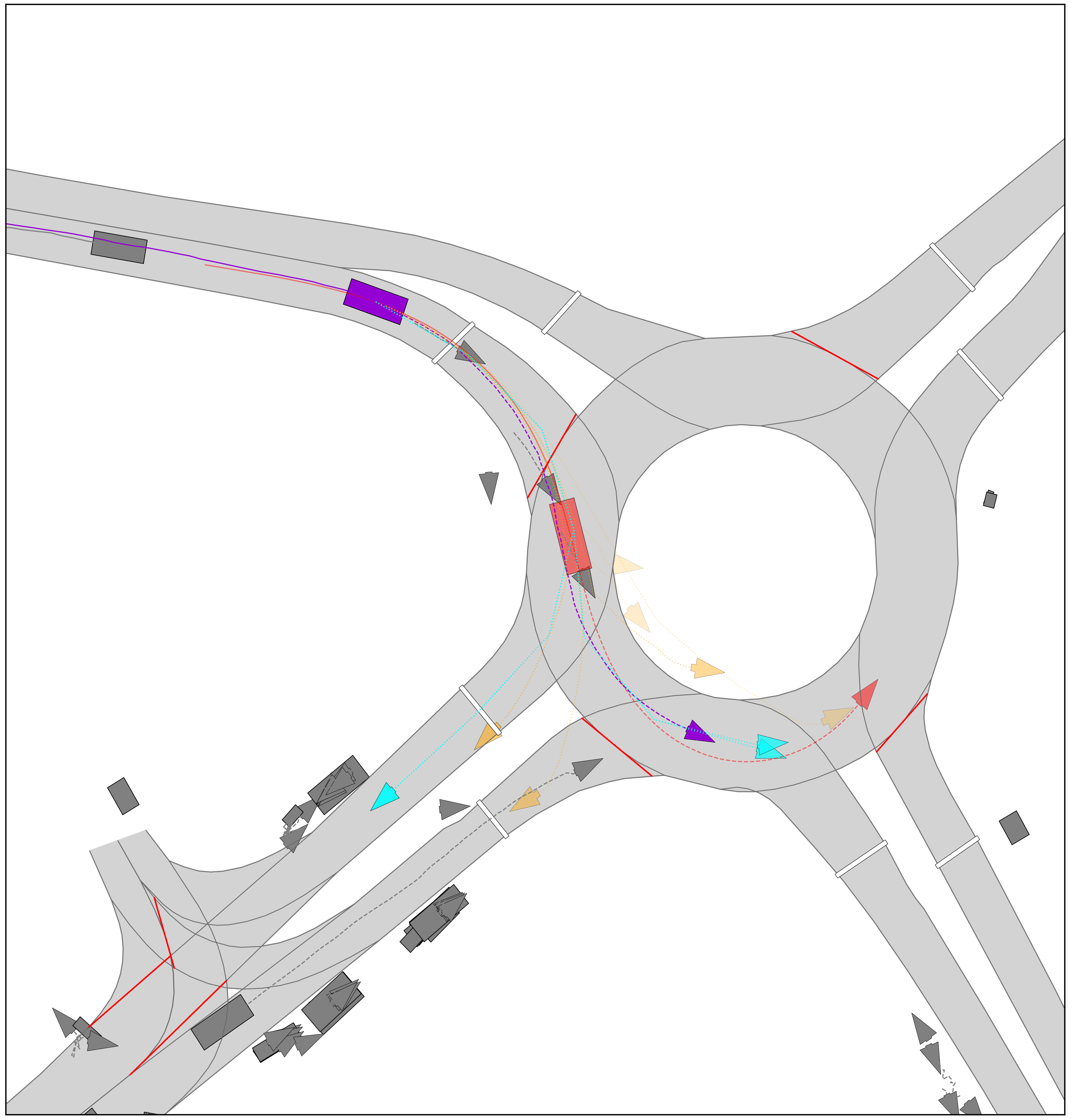}
        \label{fig:quali_argo_base_model_argo}
    }
    \caption{Comparison between models trained on L4 and AV2 in a scenario of the L4 test set. Both figures also contain trajectories a model-based approach which simply shows all legal routes (in blue).}
    \label{fig:quali_argo_base_model}
\end{figure}

\cref{fig:quali_argo_base_model} shows a roundabout scenario of our L4 test set. The model trained on the L4 dataset accurately predicts trajectories for the roundabout whereas the model trained on AV2 struggles to stay on the road. The AV2 dataset contains only a few roundabouts and is therefore far less accustomed to such situations.

\section{Conclusion}

In this work, we explore the effect of geographic diversity and variations of provided feature sets on multi-agent trajectory prediction by comparing well known datasets to our own L4 Motion Forecasting dataset. 

Our systematic evaluation of dataset design choices using the state-of-the-art QCNet model yielded three key insights. First, the inclusion of supplementary map and agent features—such as additional lane types, stop/yield lines—did not measurably improve prediction accuracy compared to baseline feature sets. This suggests that modern architectures can achieve optimal performance without relying on extensive feature engineering, validating the sufficiency of existing public datasets for capturing complex interactions. Second, cross-dataset experiments revealed significant domain shifts between datasets, particularly when transferring models trained on US-centric benchmarks (e.g., Argoverse 2) to our geographically distinct L4 dataset. However, pretraining on external datasets followed by fine-tuning on target data mitigated these challenges, highlighting the value of leveraging complementary datasets. Third, geographic diversity within training data proved critical: models trained on multi-country datasets generalized better across regions, even when data from one location (e.g., Germany) dominated the training distribution. This underscores the importance of broadening dataset scope to encapsulate varied driving cultures and infrastructure.

We see practical implications for autonomous driving research. Dataset designers should prioritize geographic and scenario diversity over feature quantity, while model developers can focus on architectures that generalize across domains rather than overfitting to localized features. Future work should explore the integration of additional regions, investigate the impact of cultural driving behaviors on prediction robustness, and develop methods to explicitly address domain shifts in trajectory forecasting.

{\small
\bibliographystyle{IEEEtran}
\bibliography{sources}
}

\end{document}